\documentclass[sigconf]{acmart}
\usepackage{comment,color}
\usepackage{hyperref}

\newcommand{\Eqref}[1]{Eq.~(\ref{#1})}
\newcommand{\Propref}[1]{Proposition~\ref{#1}}
\newcommand{\Figref}[1]{Figure~\ref{#1}}
\newcommand{\mE}{\mathbb{E}}

\newcommand{\mV}{\mathbb{V}}
\newcommand{\pdelta}{\delta^{(1)}}
\newcommand{\ndelta}{\delta^{(0)}}
\newcommand{\cvrL}{\mathcal{L}^{CVR}_{ideal}}
\newcommand{\propL}{\mathcal{L}^{Score}_{ideal}}
\newcommand{\ipsL}{\widehat{\mathcal{L}}^{CVR}_{IPS}}
\newcommand{\icvrL}{\widehat{\mathcal{L}}^{Score}_{ICVR}}
\usepackage{algorithm}
\usepackage{algpseudocode}

\AtBeginDocument{%
  \providecommand\BibTeX{{%
    \normalfont B\kern-0.5em{\scshape i\kern-0.25em b}\kern-0.8em\TeX}}}

\copyrightyear{2020} 
\acmYear{2020} 
\setcopyright{acmcopyright}\acmConference[SIGIR '20]{Proceedings of the 43rd International ACM SIGIR Conference on Research and Development in Information Retrieval}{July 25--30, 2020}{Virtual Event, China}
\acmBooktitle{Proceedings of the 43rd International ACM SIGIR Conference on Research and Development in Information Retrieval (SIGIR '20), July 25--30, 2020, Virtual Event, China}
\acmPrice{15.00}
\acmDOI{10.1145/3397271.3401282}
\acmISBN{978-1-4503-8016-4/20/07}

\begin{document}
\fancyhead{}
\title{Dual Learning Algorithm for Delayed Conversions}

\author{Yuta Saito}
\affiliation{\institution{Tokyo Institute of Technology}}
\email{saito.y.bj@m.titech.ac.jp}

\author{Gota Morishita}
\authornote{The author performed this research while employed at CyberAgent, Inc.}
\affiliation{\institution{Independent Researcher}}
\email{gota.morishita@gmail.com}

\author{Shota Yasui}
\affiliation{\institution{CyberAgent, Inc.}}
\email{yasui\_shota@cyberagent.co.jp}



\begin{abstract}
  In display advertising, predicting the conversion rate (CVR), meaning the probability that a user takes a predefined action on an advertiser's website, is a fundamental task for estimating the value of displaying an advertisement to a user. There are two main challenges in CVR prediction due to delayed feedback. First, some positive labels are not correctly observed in training data because some conversions do not occur immediately after a click. Second, delay mechanisms are not uniform among instances, meaning some positive feedback are much more frequently observed than others. It is widely acknowledged that these problems lead to severe bias in CVR prediction. To overcome these challenges, we propose two unbiased estimators: one for CVR prediction and the other for bias estimation. Subsequently, we propose a dual learning algorithm in which a CVR predictor and a bias estimator are trained in alternating fashion using only observable conversions. The proposed algorithm is the first of its kind to address the two major challenges in a theoretically sophisticated manner. Empirical evaluations using synthetic datasets demonstrate the practical value of the proposed approach.
\end{abstract}



\maketitle

\section{Introduction}
Display advertising is a way of online advertising in which advertisers pay publishers for placing ads on their websites. Over the past decade, selling display advertisements via programmatic instantaneous auction called real-time bidding has become a common practice in the display advertising domain~\cite{Muthukrishnan2009AdER}. Advertisers are offered several payment options, such as paying per click and paying per conversion (CPA). CPA has become the predominant payment method because conversions have a more direct effect on advertiser returns on investment compared to clicks. Therefore, we consider a CPA model in which advertisers pay only if a user performs a predefined conversion. A platform that supports such performance-based payment options must convert advertiser bids into an expected price per impression (eCPM) to determine the optimal bid price in an auction~\cite{chapelle2014modeling,yoshikawa2018nonparametric}. In a CPA model, eCPM depends on the conversion rate (CVR), and accurately predicting the CVR is essential for determining the optimal price to bid for each impression.

Although click-through rate prediction has been extensively studied~\cite{cheng2016wide}, it is difficult to apply these methods directly to the CVR prediction task. This is because a predictive model should be trained on fresh data to prevent data from becoming stale and to follow seasonal trends~\cite{chapelle2014modeling,ktena2019addressing}; there are two main difficulties in using fresh data for CVR prediction due to delayed feedback issue.
First, unlike a click event, a conversion does not always occur immediately after a click on an ad. While the time delay between an impression and click is usually only a few seconds, the time gap between a click and conversion can be a few hours or even days. Consequently, some conversions that will occur eventually have not yet been observed at the time of model training, and the corresponding instances are falsely considered as negative responses (\textbf{Positive-Unlabeled problem}).
The second challenge is that the missing mechanism for conversion data is missing-not-at-random (MNAR). For example, decisive users are much more likely to convert immediately after a click than indecisive users. Therefore, the probabilities of conversions being observed correctly are not uniform among samples. It is widely recognized that the MNAR mechanism can lead to sub-optimal and biased estimations (\textbf{MNAR problem})~\cite{imbens2015causal,saito2020unbiased}. 

Several works have been conducted to address the delayed feedback issue.~\cite{chapelle2014modeling} assumed that the delay distribution is exponential and proposed two models for predicting the CVR and delay distribution separately. However, this parametric assumption is often too strict for modeling complex real-world conversion data~\cite{yoshikawa2018nonparametric,ktena2019addressing}.~\cite{yoshikawa2018nonparametric} extended this study and proposed a non-parametric kernel density model for the estimation of delay distributions. However, this kernel method is considered to be unsuitable for the high-dimensional computational advertising domain because of the curse of dimensionality.~\cite{ktena2019addressing} introduced variants of the importance weighting estimator and positive-unlabeled learning as two separate approaches to solving the delayed feedback problem. However, importance weighting only tackles the MNAR problem and positive-unlabeled learning only addresses the positive-unlabeled problem. As discussed above, one has to address both the positive-unlabeled and MNAR problems for handling delayed feedback, but a method that simultaneously solves these two challenging problems has not yet been proposed.

To address the two major challenges, we first propose an unbiased estimator for the ideal loss function for CVR prediction. The proposed estimator weights each observed conversion using a parameter called the propensity score and does not make any parametric assumptions regarding the delay distribution. However, there is a difficulty in that true propensity scores are unknown in the real-world, thus they have to be estimated. To estimate propensity scores accurately, we subsequently show that the unbiased propensity estimation is possible by weighing each sample using their CVR. Based on these observations, we propose a \textit{Dual Learning Algorithm for Delayed Feedback (DLA-DF)}, which trains a CVR predictor and propensity score estimator in alternating fashion. The proposed learning framework can solve the positive-unlabeled and MNAR problems simultaneously and is expected to adjust to real-world complex delay distributions.

Finally, to evaluate the efficacy of the proposed approach in a delayed feedback setting, we conducted experiments using synthetic datasets. The results demonstrate that the proposed algorithm outperforms existing baselines, particularly in situations where delay is severe and the parametric assumptions of previously proposed methods are violated. These theoretical and empirical findings suggest that the proposed learning framework is a suitable choice for predicting CVR in realistic delayed feedback environments.

\section{Problem Setting}
Given a set of $N$ units indexed by $i$, $X_i \in \mathcal{X} \subset \mathbb{R}^d$ denotes the feature vector for each unit. Let $Y_i \in \mathcal{Y} =  \{0, 1\}$ be a random variable representing true conversion information. If an individual $i$ will eventually convert, then $Y_i = 1$. Otherwise, $Y_i = 0$. In the delayed feedback setting, true conversion variables are not fully observable because of conversion delay. To formulate such a delayed feedback setting precisely, we introduce another binary random variable $O_i \in \{0, 1\}$. This random variable represents whether or not a true outcome is observed, which depends on the elapsed time from a corresponding click. If $O_i = 1$, then a conversion is observed. Otherwise, a conversion is not correctly observed. Using these random variables, we can represent an observed outcome indicator as $Y^{obs} = O_i \cdot Y_i$. If we have observed the conversion of $i$, then $Y_i^{obs}=1$. Otherwise, $Y_i^{obs}=0$. Note that the true conversion indicator $Y_i$ is not always equal to the observed conversion indicator $Y_i^{obs}$; the conversion of $i$ is observable only when the unit will eventually convert and the true outcome is observable (i.e., $ Y_i^{obs} = 1 \Leftrightarrow O_i = 1 \& Y_i = 1 $). Finally, we use $E \in \mathbb{R}_{\ge 0}$ to denote the elapsed time since a click. When $E$ is large, the probability of a true label being correctly observed is also large. 

Throughout this paper, we assume that features affecting both $O$ and $Y$ are fully observed (i.e., $Y \perp O \, | \, X, E$), which is referred to as \textit{Unconfoundedness} in causal inference~\cite{imbens2015causal}. Building on this assumption, we obtain the following equation connecting the true CVR to the observed CVR:
\begin{align*}
    P ( Y_i^{obs} = 1 \, | \, X_i, E_i )  = \theta (X_i, E_i) \cdot \gamma (X_i)
\end{align*}
where we denote $P\left( O = 1\, | \, X, E \right) $ as $\theta (X, E)$ and $ P\left( Y = 1\, | \, X \right)$ as $\gamma(X) $. Additionally, the CVR is assumed to be independent of the elapsed time, as described in Eq. (4) in~\cite{chapelle2014modeling}.

The goal of this study is to obtain a predictor $f: \mathcal{X} \rightarrow (0, 1)$ that accurately predicts the true CVR. To achieve this goal, we define the ideal loss function that should be optimized to obtain an accurate predictor as follows:
\begin{align}
    \cvrL (f) & = \mE_{(X, Y)} \left[  Y \pdelta(f (X) ) + (1 - Y) \ndelta (f (X) )  \right] \label{eq:ideal_loss}
\end{align}
where the functions $\pdelta(\cdot) $ and $\ndelta(\cdot) $ characterize the loss function. For example, when these functions are defined as $ \pdelta(f) = - \log (f(X)), \; \ndelta(f) = - \log (1 - f(X)) $, \Eqref{eq:ideal_loss} is called binary cross entropy loss.

The loss function in \Eqref{eq:ideal_loss} is defined using the true conversion indicator, and thus, is ideal. However, in the delayed feedback setting, true conversion indicators ($Y$) are unobserved and the direct minimization of this ideal loss function is infeasible. Therefore, the critical component of the delayed feedback problem is the estimation of the ideal loss function from observable variables. 

\section{Proposed Method}

\subsection{Unbiased CVR Prediction}
To approximate the ideal loss function from observable data, here we propose an unbiased estimator for the ideal loss function for CVR prediction.

\begin{definition}(IPS estimator for the ideal loss function for CVR prediction)
\textit{When the set of propensity scores is given, the inverse propensity score (IPS) estimator for the ideal loss function is defined as}
\begin{align}
    \ipsL (f | \theta )   
     & = \frac{1}{N} \sum_{i=1}^N  \frac{Y^{obs}_i}{\theta (X_i, E_i)} \pdelta_i(f) + (1 - \frac{Y^{obs}_i}{\theta (X_i, E_i)}) \ndelta_i(f), \label{eq:ips_loss}
  \end{align}
\textit{where $ \theta (X, E) = P( O=1 \, | \, X, E ) = P(Y^{obs} = 1 \, | \, Y=1, X, E)$ is called the \textit{propensity score} and $\delta^{(\cdot)}_i (f)$ is a simplified notation for $\delta^{(\cdot)} (f (X_i) ) $.}
\end{definition}

The following proposition formally proves that the IPS estimator is statistically unbiased against the ideal loss function.
\begin{proposition}(Unbiasedness of the IPS estimator)
The IPS estimator in \Eqref{eq:ips_loss} is statistically unbiased against the ideal loss function in \Eqref{eq:ideal_loss}, i.e., $ \mE [ \ipsL (f) ] =  \cvrL (f) $.
\begin{proof}
    We can prove the unbiasedness by following the same logic flow used in Proposition 4.3 in~\cite{saito2020unbiased}.
\end{proof}
\label{proposition1}
\end{proposition}

\Propref{proposition1} validates that unbiased CVR prediction is possible by optimizing the unbiased loss function in \Eqref{eq:ips_loss} using only observable conversions.

\subsection{Unbiased Propensity Estimation}
The unbiasedness stated in \Propref{proposition1} is desirable for obtaining a CVR predictor, but is dependent on the availability of true propensity scores. In general, the estimation of propensity scores in the IPS estimator can be formulated as a classification problem. However, observation indicators are unobservable in our setting. Therefore, we propose a method for the unbiased estimation of propensity scores from observed conversions.

We first define the ideal loss function for propensity estimation as follows:
\begin{align}
\propL (g) = \mE_{(X, E, O)} \left[  O \pdelta(g (X, E) ) + (1 - O) \ndelta(g (X, E) )  \right] \label{eq:ideal_loss_score}
\end{align}
where $g: \mathcal{X} \times \mathbb{R} \rightarrow (0, 1)$ is a predictor that estimates the propensity score\footnote{$E$ is unavailable for the test data. However, the propensity score estimator is necessary only for training, and thus, the unavailability of $E$ in the test data is not an issue.}. 

We now propose an \textit{inverse conversion rate} (ICVR) estimator that shares the same structure as the IPS estimator.
\begin{definition}(ICVR estimator)
\textit{When a set of CVRs is given, the ICVR estimator for the ideal loss function in \Eqref{eq:ideal_loss_score} is defined as}
\begin{align}
    \icvrL (g | \gamma )  = \frac{1}{N} \sum_{i=1}^N  \frac{Y^{obs}_i }{\gamma (X_i)} \pdelta_i(g)  + (1 - \frac{Y^{obs}_i }{\gamma (X_i)}) \ndelta_i(g), \label{eq:icvr_loss}
  \end{align}
\textit{where $\delta^{(\cdot)}_i (g)$ is a simplified notation for $\delta^{(\cdot)} (g (X_i, E_i) ) $.}
\end{definition}

Following the same logic flow presented in \Propref{proposition1}, the next proposition proves that the ICVR estimator is statistically unbiased against the ideal loss function for propensity estimation.

\begin{proposition}(Unbiasedness of the ICVR estimator)
The ICVR estimator in \Eqref{eq:icvr_loss} is statistically unbiased against the ideal loss function in \Eqref{eq:ideal_loss_score}, i.e., $ \mE [ \icvrL (g) ] =  \propL (g) $.
\label{proposition2}
\end{proposition}

\Propref{proposition2} indicates that the unbiased propensity estimation is possible by optimizing the unbiased loss function in \Eqref{eq:icvr_loss} using only observable conversions.

\subsection{Algorithm}
Here, we describe the proposed DLA-DF algorithm, which jointly trains a propensity estimator and CVR predictor using observable conversions. 

First, given a propensity estimator $g_{\phi}$ parameterized by $\phi$, the loss function for deriving the parameters of a CVR predictor $f$ is defined as
\begin{align*}
  \ipsL \left(f \, | \, g_{\phi} \right)  = \frac{1}{N} \sum_{i=1}^N  \frac{Y^{obs}_i}{g_{\phi} (X_i,E_i)} \pdelta_i(f)  + (1 - \frac{Y^{obs}_i}{g_{\phi} (X_i,E_i)}) \ndelta_i(f)
\end{align*}

Next, given a CVR predictor $f_{\psi}$ parameterized by $\psi$, the loss function for deriving the parameters of a propensity estimator $g$ is defined as
\begin{align*}
   \icvrL \left(g \, | \, f_{\psi} \right) = \frac{1}{N} \sum_{i=1}^N  \frac{Y^{obs}_i}{f_{\psi} (X_i)} \pdelta_i(g) 
   + (1 - \frac{Y^{obs}_i}{f_{\psi} (X_i)}) \ndelta_i(g) 
\end{align*} 

The detailed procedure for the proposed DLA-DF algorithm is described in Algorithm 1.

\begin{algorithm}[t]
\caption{Dual Learning Algorithm for Delayed Feedback}
\begin{algorithmic}[1]
\Require{training data $\mathcal{D} = \{X_i,E_i, Y_i^{obs}\}_{i=1}^N$; mini-batch size $m$; learning rate $\eta$}
\Ensure{model parameters $\psi$ and $\phi$}
\State Initialize parameters with random weights $\psi$, $\phi$
\Repeat
  \State{Sample mini-batch data $\{X_j, E_j, Y^{obs}_j \}_{j=1}^{m}$ from $\mathcal{D}$}
    \State Update $\psi$ by $ \nabla_{\psi}  \ipsL (f_{\psi} \, | \, g_{\phi} ) $ with a fixed $\phi$
    \State Update $\phi$ by $\nabla_{\phi}  \icvrL (g_{\phi} \, | \, f_{\psi} )$ with a fixed $\psi$
\Until convergence;
\State \Return $\psi, \phi$
\end{algorithmic}
\end{algorithm}

\subsection{Variance Reduction Technique}
The proposed learning framework is theoretically refined and promising, however, unbiased estimators derived using inverse propensity weighting are widely known to exhibit large variance~\cite{saito2020unbiased}. Therefore, we analyze the variance of the unbiased estimators and propose a method for addressing this variance issue.
\begin{theorem}(Variance of the unbiased estimators) Given sets of independent random variables $ \{ ( Y^{obs}_i,  O_i, Y_i )\}$, propensity scores $\{ \theta (X_i, E_i) \}$, and a CVR predictor $f$, the variance of the IPS estimator is
\begin{align*}
    \mV \left( \ipsL (f) \right)  = \frac{1}{ N^2 } \sum_{i=1}^N  \gamma(X_i) \left( \frac{1}{\theta (X_i, E_i)} - \gamma(X_i) \right) \left(\pdelta_i (f) - \ndelta_i (f)  \right)^2.
 \end{align*}
 Replacing $\gamma, \theta, \pdelta_i (f) $, and $\ndelta_i (f) $ with $\theta, \gamma, \pdelta_i (g) $, and $\ndelta_i (g) $ yields the variance of the ICVR estimator.
 \begin{proof}
     We can derive the variance by following the same logic flow presented in Theorem 4.4 of~\cite{saito2020unbiased}.
 \end{proof}
\end{theorem}

\begin{figure*}[ht]
    \centering
    \begin{center}
        \begin{tabular}{c}
            \begin{minipage}{0.325\hsize}
                \begin{center}
                    \includegraphics[clip, width=5.7cm]{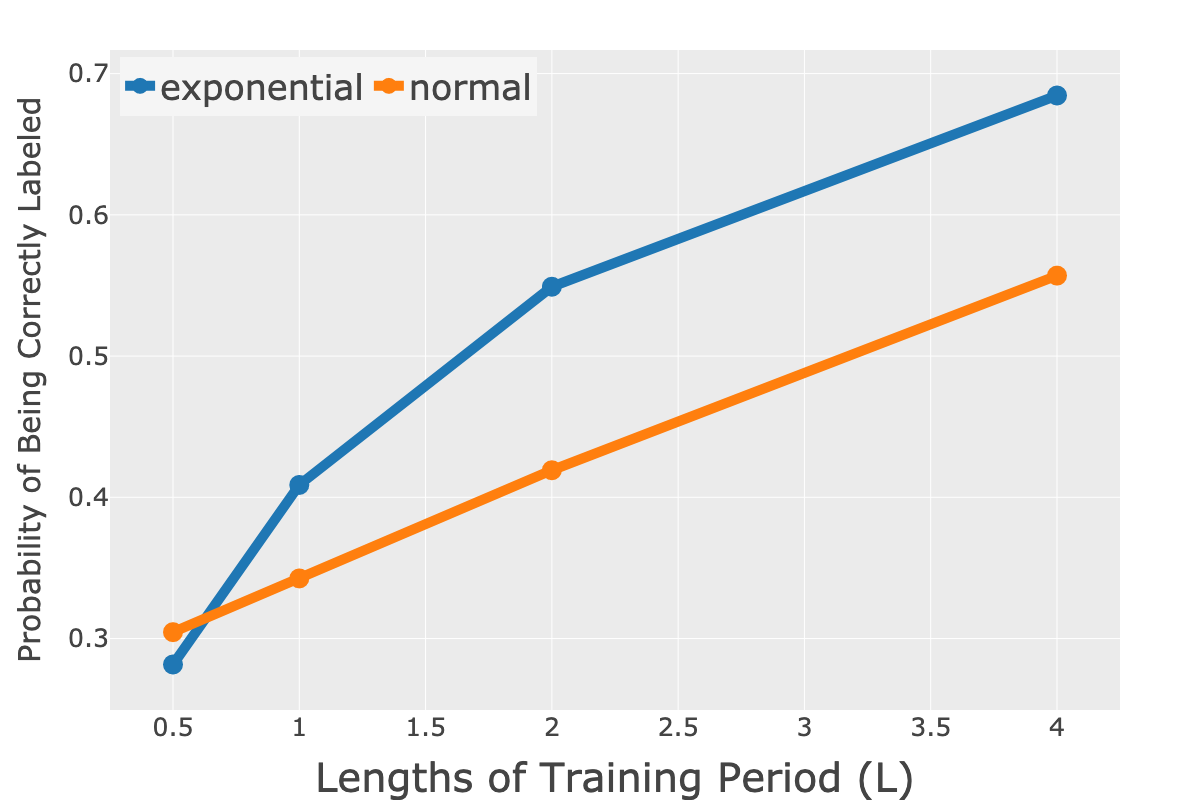}
                \end{center}
            \end{minipage}
            
            \begin{minipage}{0.325\hsize}
                \begin{center}
                    \includegraphics[clip, width=5.7cm]{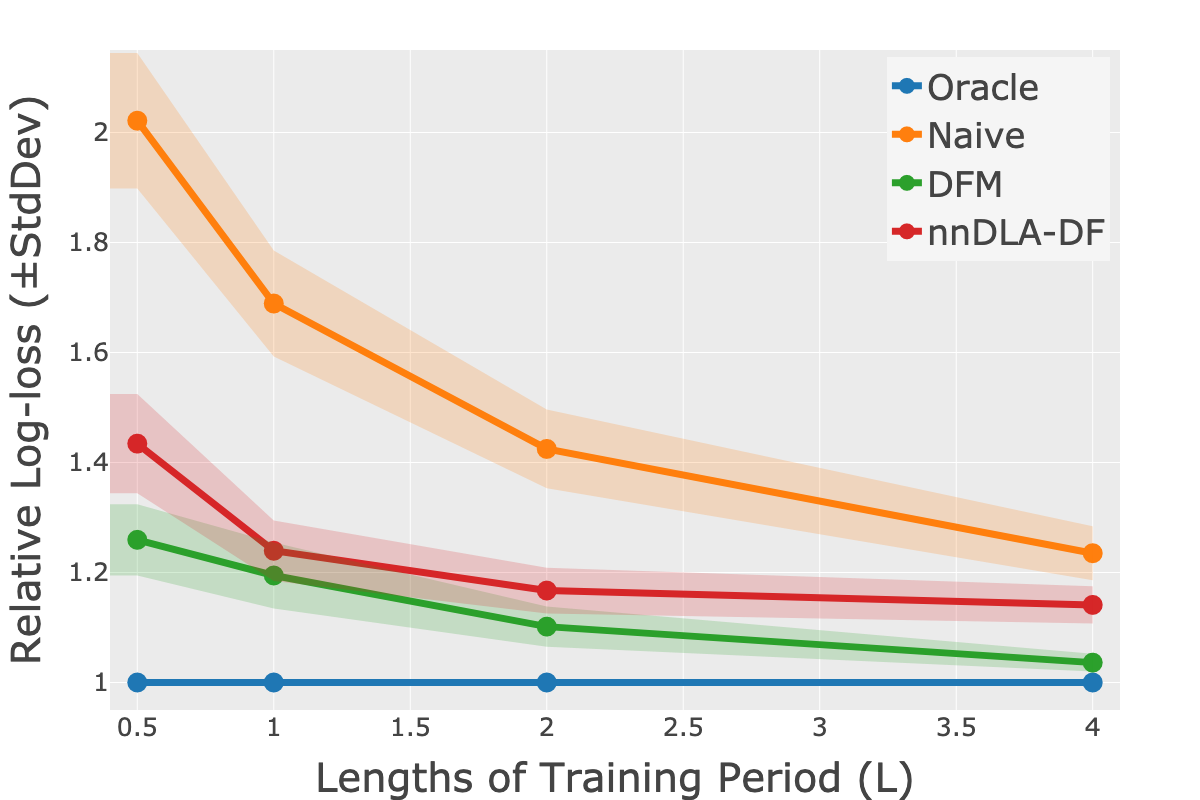}
                \end{center}
            \end{minipage}
            
            \begin{minipage}{0.325\hsize}
                \begin{center}
                    \includegraphics[clip, width=5.7cm]{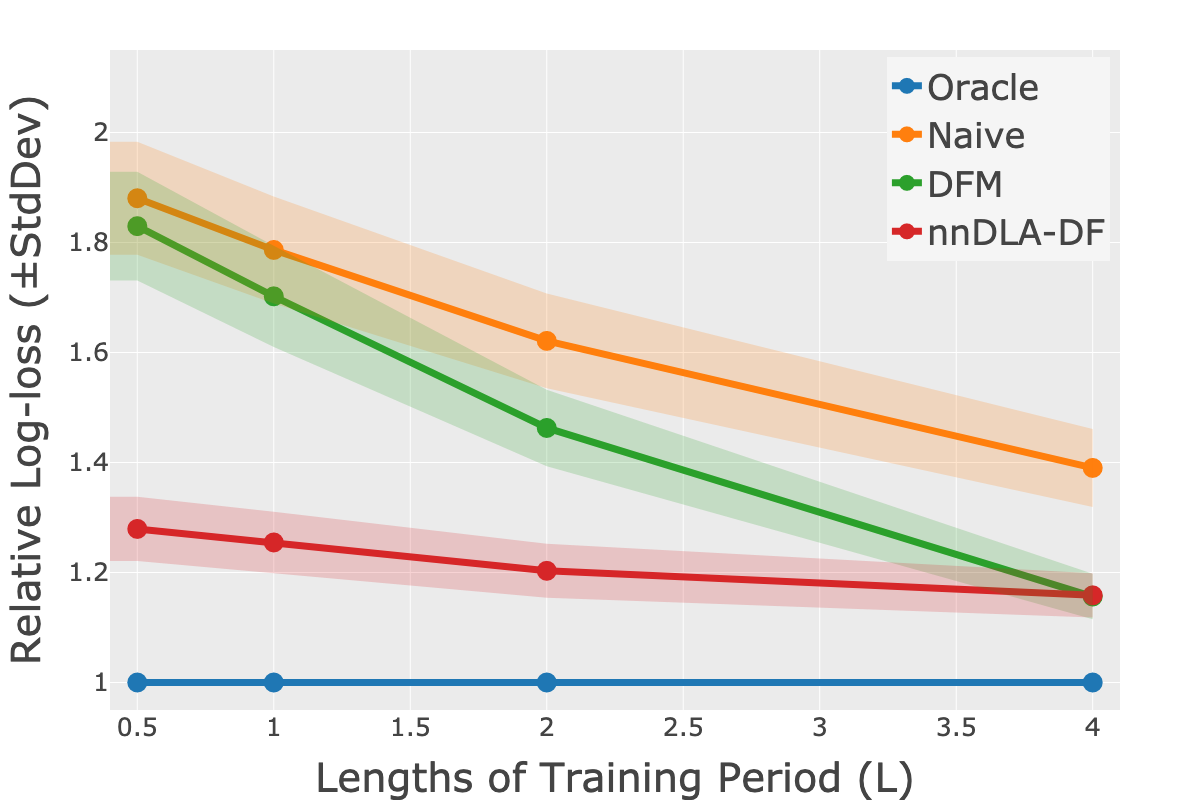}
                \end{center}
            \end{minipage} 
        \end{tabular}
    \end{center}
    \caption{(left) Averaged propensity score for each length of training period. (center) Relative log-loss on test sets when the delay distribution is exponential. (right) Relative log-loss on test sets when the delay distribution is normal.}
    \label{fig:results}
\end{figure*}

The variance depends on the inverse of the propensity scores, and thus, can be huge, particularly when severe delay occurs. Therefore, we propose utilizing the following \textit{non-negative estimator}~\cite{kiryo2017positive} to address the variance problem of the unbiased estimators.
\begin{definition}\textit{(Non-negative estimator) When propensity scores are given, the non-negative estimator is defined as}
\begin{align}
  \widehat{\mathcal{L}}_{\textit{non-negative}} \left( f \right)  = \frac{1}{N} \sum_{i=1}^N \max \left\{ \ell_{IPS} \left( f \right), \, 0 \right\}, \label{eq:non_negative_loss}
\end{align}
\textit{where $ \ell_{IPS} \left( f \right) = \frac{Y^{obs}_i}{\theta (X_i, E_i)} \pdelta_i(f)  + (1 - \frac{Y^{obs}_i}{\theta (X_i, E_i)}) \ndelta_i(f) $. The nonnegative variant for the ICVR estimator can be defined similarly. This non-negative estimator provides a lower variance than the IPS and ICVR estimators at the cost of introducing some bias.}
\end{definition}

\begin{algorithm}[t]
\caption{Synthetic data generation process}
\begin{algorithmic}[1]
\Require{The number of click events $N$, number of features observed for each event $p$, length of training periods $L$, and standard deviations of the distributions $\sigma_X$ and $\sigma_W$.}
\Ensure{dataset $\{X_i, D_i, E_i, Y_i^{obs} \}_{i=1}^n$.}
\State Sample true coefficient vectors: $\boldsymbol{W}_{cvr}, \boldsymbol{W}_{expo}  \sim \mathcal{N}(\bold{0}_p, \sigma^2_W \boldsymbol{I}_p)$
\For{$i=1, \ldots, n$}
\State Sample feature vectors: $X_i \sim \mathcal{N} ( \bold{0}_p, \sigma^2_X \boldsymbol{I}_p )
$.
\State Calculate true CVRs: $\gamma(X_i) = \text{sigmoid} \left( \boldsymbol{W}_{cvr} X_i \right)$
\State Sample time stamps of click: $ts\_click_i \sim Unif(0, L)$
\State Sample the lengths of delay: $ D_i \sim \mathcal{D}_{delay}(\text{exp} \left( \boldsymbol{W}_{expo} X_i \right)) $
\State Calculate the elapsed time from the click: $E_i = L - ts\_click_i$ 
\State Calculate the true observation variables: $ O_i = \mathbb{I} \{ D_i \le L \} $
\State Sample the true conversion indicators: $ Y_i \sim Bern (\gamma(X_i)) $
\State Calculate the observed conversion indicators: $Y_i^{\text{obs}} = O_i \cdot Y_i$
\EndFor
\State \Return $\{X_i, D_i, E_i, Y_i^{obs} \}_{i=1}^n$
\end{algorithmic}
\end{algorithm}

\section{Synthetic Experiment}
In this section, we present an empirical comparison of the proposed method to baseline methods using a synthetic dataset.

\subsection{Experimental Setup}

\subsubsection{Synthetic data generation procedure}
We created a synthetic dataset simulating a delayed feedback setting. The data generation procedure is presented in Algorithm 2\footnote{$\text{sigmoid}(\cdot)$ is the sigmoid function, $Unif(\cdot, \cdot)$ is a uniform distribution, and $Bern(\cdot)$ is a Bernoulli distribution.}. We set $N=100,000$, $p=30$, $\sigma_X = 0.5$, and $\sigma_W=1.0$. For the delay distribution ($\mathcal{D}_{delay}$), we considered exponential and normal distributions. The lengths of the training period $L$ were set to $0.5, 1, 2, 4$ (days). A smaller value of $L$ yields smaller propensities, as indicated in Figure 1 (left).

\subsubsection{Baselines and the proposed method}
We compared the performances of the following methods. 
\textbf{Oracle}: A logistic regression model trained using true conversion data ($Y$), which is unobservable in the real-world. Therefore, the performance of the oracle model is the best achievable prediction performance. 
\textbf{Naive}: A logistic regression model trained naively using observed conversions ($Y^{obs}$). 
\textbf{Delayed Feedback Model (DFM)~\cite{chapelle2014modeling}}: This model is a widely used baseline in delayed conversion settings~\cite{yoshikawa2018nonparametric,ktena2019addressing} and assumes that the delay distribution is exponential.
\textbf{Non-negative Dual Learning Algorithm (nnDLA-DF)}: This is the proposed method. We used a logistic regression model for both the CVR predictor ($f$) and propensity estimator ($g$). Both estimators were trained using non-negative loss function in \Eqref{eq:non_negative_loss}.

\subsection{Results}
\Figref{fig:results} (center) and (right) present the values of the log-loss on the test sets relative to the performance of the oracle model when the delay follows exponential and normal distributions, respectively. For both figures, averaged relative log-loss on test sets and its standard deviations over 10 iterations are reported.

\Figref{fig:results} (center) demonstrates that the proposed nnDLA-DF is outperformed by DFM. This result is reasonable because the DFM's assumption of an exponential delay distribution is perfectly satisfied in this setting. However, the proposed method exhibits competitive and stable performance, despite that it does not assume any assumption regarding the delay distribution. \Figref{fig:results} (right) demonstrates that the proposed method significantly outperforms the other methods when $L=0.5,1,2$. In contrast, the benefits of DLA-DF are much smaller when $L=4$, but it is not outperformed by the other methods in any setting, which demonstrates the stable prediction performance of the proposed algorithm.


\section{Conclusion}
In this study, we explored the delayed feedback problem, where true conversion indicators are not fully observable due to conversion delay. To address this problem, we developed the DLA-DF algorithm, which is the first to solve both the positive-unlabeled and MNAR problems of delayed conversions simultaneously. Additionally, the proposed framework does not depend on any parametric assumptions regarding delay distributions and is able to perform well in a wide range of situations. In empirical evaluations, the proposed algorithm outperformed existing baselines, particularly in practical settings where there exists severe delay or the parametric assumptions regarding delay distributions are no longer satisfied.


\end{document}